\documentclass[runningheads]{llncs}
\usepackage{graphicx}
\usepackage{array}
\usepackage{tabu}
\usepackage[caption=false]{subfig}
\usepackage{multirow}

\usepackage{floatrow}
\newfloatcommand{capbtabbox}{table}[][\FBwidth]

\newcolumntype{M}[1]{>{\centering\arraybackslash}m{#1}}
\newcolumntype{L}[1]{>{\arraybackslash}m{#1}}

\usepackage{float}
\floatstyle{plaintop}
\restylefloat{table}

\begin{document}

\title{Results of the seventh edition of the BioASQ Challenge}
%
%

\author{
Anastasios Nentidis$^{1,2}$ \and
Konstantinos Bougiatiotis$^{1,3}$ \and
Anastasia Krithara$^1$ \and 
Georgios Paliouras$^{1,4}$}

\authorrunning{A. Nentidis et al.}
%
\institute{
$^1$National Center for Scientific Research ``Demokritos'', Athens, Greece \\
\email{\{tasosnent,  bogas.ko, akrithara, paliourg\}@iit.demokritos.gr}\\
  $^2$Aristotle University of Thessaloniki, Thessaloniki, Greece\\
  $^3$National and Kapodistrian University of Athens, Athens, Greece\\
  $^4$University of Houston, Texas, USA
}

\maketitle              
\begin{abstract}
The results of the seventh edition of the BioASQ challenge are presented in this paper. The aim of the BioASQ challenge is the promotion of systems and methodologies through the organization of a challenge on the tasks of large-scale biomedical semantic indexing and question answering. In total, 30 teams with more than 100 systems participated in the challenge this year. As in previous years, the best systems were able to outperform the strong baselines. This suggests that state-of-the-art systems are continuously improving, pushing the frontier of research. \\

\textbf{\textcopyright2019}. This manuscript version is made available under the CC-BY-NC-ND 4.0 license http://creativecommons.org/licenses/by-nc-nd/4.0/

\end{abstract}
\section{Introduction}

The aim of this paper is twofold. First, we aim
to give an overview of the data issued during the
BioASQ challenge in 2019. In addition, we aim to
present the systems that participated in the challenge
and evaluate their performance. To
achieve these goals, we begin by giving a brief overview of the tasks, which took place from February to May 2019, and the challenge's data. Thereafter, we provide an overview of the systems that participated
in the challenge. Detailed descriptions of some of the systems are
given in the workshop proceedings. The evaluation of the systems, which was carried out  using state-of-the-art measures or manual assessment, is the last focal point of this paper, with remarks regarding the results of each task. The conclusions sum up this year's challenge.

\section{Overview of the Tasks}

The challenge comprised two tasks: (1) a large-scale
biomedical semantic indexing task (Task 7a) and (2) a biomedical question answering task (Task 7b). In this section a brief description of the tasks is provided focusing on differences from previous years and updated statistics about the corresponding datasets. A complete overview of the tasks and the challenge is presented in \cite{Tsatsaronis2015}.

\subsection{Large-scale semantic indexing - 7a}
In Task 7a the goal is to classify documents from the PubMed digital library into concepts of the MeSH hierarchy. Here, new PubMed articles that are not yet annotated by MEDLINE indexers are collected and used as test sets for the evaluation of the participating systems. Similarly to task 5a and 6a, articles from all journals were included in the test data sets of task 7a. As soon as the annotations
are available from the MEDLINE indexers, the performance of each system is calculated using standard flat information retrieval measures, as well as, hierarchical ones.  As in previous years, an on-line and large-scale scenario was provided, dividing the task into three independent batches of 5 weekly test sets each. Participants had 21 hours to provide their answers for each test set. 
Table \ref{tab:7a_data} shows the number of articles in each test
set of each batch of the challenge. 14,200,259 articles with 12.69 labels per article, on average, were provided as training data to the participants.

\begin{table}[!t]
        \centering
        \begin{tabular}{M{0.12\linewidth}M{0.18\linewidth}M{0.21\linewidth}M{0.21\linewidth}}\hline
        \textbf{Batch} & \textbf{Articles} & \textbf{Annotated Articles} & \textbf{Labels per Article}  \\ \hline
          \multirow{5}{*}{1} &7,358	&7,194	&	11.67\\&7,166	&7,021	& 12.95\\&11,019	&10,831	&	13.04\\&5,566	&5,482	&	12.32\\&6,729	&6,353	&	12.96\\ \cline{2-4}
        \textbf{Total} & 37,838&	36,881&	12.31 \\ \hline
        \multirow{5}{*}{2} &6,380	&6,098&12.51\\&6,785	&6,621&12.75\\&6,207	&5,927&12.75\\&7,382	&7,079&13.00\\&7,240	&6,756&12.65\\\cline{2-4}
        \textbf{Total} &33,994	&32,481&12.27\\ \hline
        \multirow{5}{*}{3} &6,266&	5,835&	12.58\\&11,455& 10,386&	12.86\\&4,750&	3,947&	12.67\\&7,338&5,021&	12.70\\&6,920&4,554&	12.63\\ \cline{2-4}
        \textbf{Total} &36,729& 29,743& 12.14 \\ \hline
        \hline
        \end{tabular}
        \caption{Statistics on test datasets for Task 7a.}\label{tab:7a_data}
\end{table}

\subsection{Biomedical semantic QA - 7b}

The goal of Task 7b was to provide a large-scale question answering challenge where the systems had to cope with all stages of a question answering task for four types of biomedical questions: “yes/no”, “factoid”, “list” and “summary” questions \cite{balikas13}.
As in previous years, the task comprised two phases: In phase A, BioASQ released 100 questions and participants were asked to respond with relevant elements from specific resources, including relevant MEDLINE articles, relevant snippets extracted from the articles, relevant concepts and relevant RDF triples. 
In phase B, the released questions were enhanced with relevant articles and snippets selected manually and the participants had to respond with \textit{exact answers}, as well as with summaries in natural language (dubbed \textit{ideal answers}).
The task was split into five independent batches and the two phases for each batch were run with a time gap of 24 hours. In each phase, the participants received 100 questions and had 24 hours to submit their answers. 
Table \ref{tab:7b_data} presents the statistics of the
training and test data provided to the participants.
The evaluation included five test batches. 

\begin{table}[!ht]
        \centering
        \begin{tabular}{M{0.18\linewidth}M{0.1\linewidth}M{0.23\linewidth}M{0.23\linewidth}}\hline
        \textbf{Batch} & \textbf{Size} & \textbf{Documents} & \textbf{Snippets}  \\ \hline
        Train&	2,747&	11.14&	13.91\\
        Test 1&	100&	3.07&	3.93\\
        Test 2&	100&	2.64&	3.22\\
        Test 3&	100&	3.08&	4.05\\ 
        Test 4&	100&	2.78&	3.71\\
        Test 5&	100&	2.39&	2.62\\
        \textbf{Total}&	3,247&	9.85&	12.31\\
        \hline
        \end{tabular}
        \caption{Statistics on the training and test datasets of Task 7b. All the numbers for the documents and snippets refer to averages.}\label{tab:7b_data}
\end{table}

\section{Overview of Participants}
\subsection{Task 7a}
For this task, 12 teams participated and results from 30 different systems were submitted. In the following paragraphs we describe those systems for which a description was available,
stressing their key characteristics. An overview of the systems and their approaches can be seen in Table \ref{tab:7a_sys}.

\begin{table}[!htb]
        \centering
        \begin{tabular}{M{0.3\linewidth}M{0.5\linewidth}}\hline
        \textbf{System} & \textbf{Approach} \\ \hline
        ceb & CNN, embeddings, ensembles\\\hline
        DeepMesh & d2v, tf-idf, MESHlabeler, attention scheme, PLT \\\hline
        Iria & bigrams, Luchene Index, k-NN, ensembles, UIMA ConceptMapper\\\hline
        MeSHProbeNet-P & Bidirectional RNN (GRU), attention scheme, encoder-decoder architecture\\\hline
        Semantic NoSQL KE & UIMA ConceptMapper, par2vec, DeepLearning4j\footnote{https://deeplearning4j.org/ Accessed June 2019}\\\hline
        
        \end{tabular}
         \caption{Systems and approaches for Task 7a. Systems for which no description was available at the time of writing are omitted. }\label{tab:7a_sys}
\end{table}

The National Library of Medicine (NLM) team, in its ``\textit{ceb}'' systems \cite{Rae2019}, adopts an end-to-end deep learning  architecture with Convolutional Neural Networks (CNN) \cite{liu2017deep} to improve the results of the Medical Text Indexer (MTI) \cite{morkBioasq2014}. In particular, they combine text embeddings with journal information. They also consider information about the years of publication and indexing, to capture concept drift and variations in the MeSH vocabulary respectively. They also experiment with an ensemble of independently trained DL models. 

The Fudan University team builds upon their previous ``\textit{DeepMeSH}'' systems, which are based on document to vector (\textit{d2v}) and tf-idf feature embeddings \cite{peng2016}, the MESHLabeler system \cite{liu2015} and learning to rank (LTR). This year, they incorporate AttentionXML \cite{You2018}, a deep-learning-based extreme multi-label text classification model, in the ``\textit{DeepMeSH}''framework. In particular, AttentionXML combines a multi-label attention mechanism, to capture label-specific information, with a shallow and wide probabilistic label tree (PLT) \cite{Jain2016}, for improved efficiency.

The ``\textit{Iria}'' systems \cite{ribadascole} are based on the same techniques used by their systems for the previous version of the challenge which are summarized in Table \ref{tab:7a_sys} and described in the corresponding challenge overview \cite{nentidis2017results}. 

The ``\textit{MeSHProbeNet-P}'' systems are upgraded versions of MeSHProbeNet \cite{Xun2019}, which participated in BioASQ6 with the name ``\textit{xgx}''. Their approach is based on an end-to-end deep learning model with an encoder-decoder architecture. The encoder consists of a recurrent neural network with multiple attentive MeSH probes to extract different aspects of biomedical knowledge from each input article.  In ``\textit{MeSHProbeNet-P}'' the attentive MeSH probes are also personalized for each biomedical article, based on the domain of each article as expressed by the journal where it has been published. 

Finally, the ``\textit{Semantic NoSQL KE}'' system variants \cite{Bernd2019} were developed extending previous year's ``\textit{SNOKE}'' systems. The systems are based on the ZB MED Knowledge Environment \cite{ZBMed2017}, utilizing the Snowball Stemmer \cite{snowball:2000} and the UIMA \cite{tanenblatt2010} ConceptMapper to find matches between MeSH terms and words in the title and abstract of each target document, adopting different matching strategies. Paragraph Vectors \cite{Le2014}  trained on the BioASQ corpus are used to rank and filter all the MeSH headings suggested by the UIMA-based framework for each document.

Similarly to the previous year, two systems developed by NLM to assist the indexers in the annotation of MEDLINE articles, served as baselines for the semantic indexing task of the challenge. MTI \cite{morkBioasq2014} with some enchantments introduced in \cite{zavorin2016} and an extension of it, incorporating features of the winning system of the first BioASQ challenge \cite{tsoumakasBioasq}.

\subsection{Task 7b}

The question answering task was tackled by 73 different systems, developed by 18 teams. In the first phase, which concerns the retrieval of information required to answer a question, 6 teams with 23 systems participated. In the second phase, where teams are requested
to submit exact and ideal answers, 13 teams with 52 different systems participated. 
An overview of the technologies employed by each team can be seen in Table \ref{tab:7b_sys}.

\begin{table}[!htb]
        \centering
        \begin{tabular}{M{0.2\linewidth}M{0.1\linewidth}M{0.5\linewidth}}\hline
        \textbf{Systems} & \textbf{Phase}& \textbf{Approach} \\ \hline
        AUTH & A, B & MetaMap, BeCAS, Lucene Index, ElasticSearch, Wordnet, ELMo, SentiWordnet, w2vec, BiLSTM \\\hline 
        AUEB & A & BM25, w2vec, BERT, DL (BCNN, PACRR, PDRMM) \\ \hline
        MindLab & A & ElasticSearch, BM25, QuickUMLS, w2vec, WMD, DL (CNN)  \\\hline 
        \_sys & A & Word and Sentence embeddings, Pseudo Relevance Feedback, BM25, LSI\\\hline
        BJUTNLP & B & SQUAD, GloVe, BiLSTM, Pointer Network \\\hline
        BIOASQ\_VK & B & ELMo, DMN attention mechanisms,  NLTK-VADER \\\hline
        DMIS & B & BioBERT, SQUAD, transfer learning \\\hline
        google & B & BERT, CoQA, Natural Questions \\\hline
        L2PS & B & SQUAD, Quasar-T, DRQA (RNN, LSTM), PSPR (LSTM), BioBERT \\\hline
        LabZhu & B & PubTator, Stanford POS tool, SPARQL\\\hline      
        MQU & B & w2vec, tf-idf, DL (LSTM), Reinforcement Learning \\\hline
        UNCC & B & BioBERT, SQUAD, Stanford POS tool, AllenNLP entailment \\\hline
        unipi-quokka-QA & B & ELMo, ELMo-PubMed, BERT, BioBERT, SciSpacy  \\\hline
        \hline
        \end{tabular}
        \caption{Systems and approaches for Task7b. Systems for which no information was available at the time of writing are omitted.}\label{tab:7b_sys}
\end{table}


The ``\textit{AUTH}'' team participated in both phases of Task 7B, with focus on phase B. For the document retrieval task, they experimented with approaches based on the BioASQ search services and ElasticSearch, querying with the conjunction of words in each question for the top 10 documents. In Phase B, for factoid and list questions they used updated versions of their BioASQ6 system \cite{Dimitriadis2019}, based on word embeddings, MetaMap \cite{AronsonL10}, BeCAS \cite{Nunes2013} and WordNet. For yes/no questions they experiment with different deep learning methods, based on ELMo embeddings \cite{Peters2018}, SentiWordnet \cite{Esuli06sentiwordnet:a} and similarity matrices to represent the question/answer pairs and use them as input for different BiLSTM architectures \cite{Dimitriadis2019bioasq}. 


The ``\textit{AUEB}'' team participated in Phase A on document and snippet retrieval tasks yielding great results. They built upon their BioASQ6 document retrieval systems \cite{Brokos2018,mcdonald2018}, which they modify to yield a relevance score for each sentence and experiment with BERT and PACRR \cite{mcdonald2018} for this task. For snippet retrieval, they utilize a BCNN \cite{yin2016abcnn} model and a model based on POSIT-DRMM (PDRMM) \cite{mcdonald2018}.
They also introduce JPDRMM, a novel deep learning approach for joint document and snippet ranking, based on PDRMM \cite{Pappas2019}.

Another approach based on deep learning methodologies for Phase A, focusing again on document and snippet retrieval, was proposed by the ``\textit{MindLaB}'' team from the National University of Colombia \cite{Vargas2019}. For the document retrieval they use the BM25 model \cite{robertson:1976} and ElasticSearch \cite{gormley2015elasticsearch} for efficiency, along with a Word Mover's Distance \cite{kusner2015word} based re-ranking scheme. For snippet retrieval, as in the previous approach, they utilized a very large collection of PubMed articles to train a CNN with similarity matrices of question-answer pairs. More specifically, they employ the BioNLPLab\footnote{http://bio.nlplab.org Accessed June 2019} w2vec embeddings that take into account the Part of Speech of each word. Also, they deploy the QuickUMLS \cite{soldaini2016quickumls} tool to create a cui2vec embedding for each snippet. 

The ``\textit{\_sys}'' systems also participated in Phase A of Task 7B. These systems filter the queries, using stop-word lists and regular expressions, and expand them using word embeddings and pseudo-relevance feedback. Relevant documents are retrieved, utilizing Query Likelihood with bigrams and BM25, and reranked, based on Latent Semantic Indexing (LSI) and document vectors. In particular, document vectors based on averaging sentence embeddings are adopted. Finally, different lists of documents are merged to form the final result, considering the position of the documents in each list.


In phase B, most systems focused on using embeddings and deep learning methodologies to tackle the tasks. For example the ``\textit{BJUTNLP}'' system utilizes the SQUAD Dataset for pre-training. The system uses both GloVe embeddings \cite{pennington2014glove} (fine tuned during training) and character-level word embeddings (through a 1-dimensional CNN) as input to a BiLSTM model and for each question a Pointer Network \cite{see2017get} is finally responsible for pinpointing the exact start and end position of the answer in the relevant snippets.

The ``\textit{BIOASQ\_VK}'' systems were based on BioBERT \cite{lee2019biobert}, but with novel modifications to allow the model to cope with yes/no, factoid and list questions \cite{Kanjirangat2019}. They pre-trained the model on the SQUAD dataset (for factoid and list questions) and SQUAD2 (for yes/no questions) to leverage the small size of the BioASQ dataset and by exploiting different pre-/post-processing techniques they obtained great results on all subtasks.

The ``\textit{DMIS}'' systems focused on the importance of the information (words, phrases and sentences) for a given question \cite{Yoon2019}. To this end, sentence level embeddings based on ELMo embeddings \cite{Peters2018} and attention mechanisms facilitated by Dynamic Memory Networks (DMN) \cite{kumar2016ask} are deployed. Moreover, sentiment analysis is performed on yes/no questions to guide the classification (positive corresponds to yes) using the NLTK-VADER \cite{hutto2014vader} tool.

The ``\textit{google}'' systems \cite{Hosein2019}, focus on factoid questions and are based on BERT based models \cite{devlin2018bert}, specifically the one in \cite{alberti2019bert} trained on the Natural Questions \cite{kwiatkowski2019natural} dataset, while also utilizing the CoQA \cite{reddy2019coqa} and the BioASQ datasets. They experiment with different input to the models, including the abstracts of relevant articles, the provided gold snippets and predicted relevant snippets. In particular, they focus on error propagation in end-to-end information retrieval and question answering systems, reaching the interesting conclusion that the information retrieval part is a bottleneck for such end-to-end QA systems.

 Interesting results come from the ``\textit{L2PS}'' team where they quantify the importance of pre-training and fine-tuning models for question answering and view the task under different regimes, namely Reading Comprehension (RC) and Open QA \cite{Kamath2019}. For the RC regime they use DRQA's document reader \cite{chen2017reading} while for the Open QA they utilize the  PSPR model \cite{lin2018denoising}. They experiment with different datasets (SQUAD \cite{rajpurkar2016squad} for RC and Quasar-T \cite{dhingra2017quasar} for Open QA) for fine-tuning the models, as well as BioBert \cite{lee2019biobert} embeddings to gain insights on the effect of the context length in this task.

The ``\textit{LabZhu}'' \cite{zhang2015} systems improved upon their systems from BioASQ6, with focus on exact answer generation. In particular, for factoid and list questions they developed two distinct approaches. One based on traditional information retrieval approaches, involving candidate answer generation and ranking, and one Knowledge-Graph based approach. In the latter approach, the answer type and the topic entity of the question are predicted and a SPARQL query is generated based on them and used to retrieve some results from the Knowledge Graph. Finally, the results of the two approaches are combined for the final answer of the question. 

The Macquarie University (``\textit{MQU}'') team focused on ideal answers and approached the task under a classification approach for snippet relevance \cite{Molla2019}. Extending their previous work \cite{Diego2017,molla2018macquarie} the snippets are marked as summary relevant or not, utilizing w2vec embeddings and tf-idf vectors of the question-sentence pairs, showcasing that a classification scheme is more appropriate than a regression one. Also, based on their previous work \cite{molla_REINFORCE:2017}, they conduct experiments using reinforcement learning towards the ROUGE score of the ideal answers and a correlation analysis between various ROUGE metrics and the BioASQ human evaluation scores, observing poor correlation of the ROUGE-Recall score with human evaluation.
 
 The ``\textit{UNCC}'' team focused on factoid, list and yes/no questions \cite{Telukuntla2019}. Their work is based on the BioBERT \cite{lee2019biobert} embeddings fine-tuned on previous years of BioASQ. They also utilize the SQUAD dataset for factoid answers and incorporated the Lexical Answer Type (LAT) \cite{ferrucci2010building} and POS-tags along with hand made rules to address specific errors of the system. Furthermore, they incorporated the entailment of the candidate sentences in yes/no questions using the AllenNLP library\cite{Gardner2017AllenNLP}.

Finally, the ``\textit{unipi-quokka-QA}'' system tackled all the different question types in phase B \cite{Resta2019}. Their work focused on experimenting with different Transformer models and embeddings, namely: ELMo, ELMo-Pumbed, BERT and BioBERT. They used different strategies depending on the question type, such as ensembles on yes/no questions, biomedical named entity extraction (using SciSpacy \cite{Neumann2019ScispaCyFA}) on list questions and different pre-/post-processing procedures.

In this challenge too, the open source OAQA system proposed by \cite{yang2016learning} served as baseline for phase B. The system which achieved among the highest performances in previous versions of the challenge remains a strong baseline for the exact answer generation task. The system is developed based on the UIMA framework. ClearNLP is employed for question and snippet parsing. MetaMap, TmTool \cite{Wei2016}, C-Value and LingPipe \cite{baldwin2003lingpipe} are used for concept identification and UMLS Terminology Services (UTS) for concept retrieval. The final steps include identification of concept, document and snippet relevance, based on classifier components and scoring, ranking and reranking techniques.

\section{Results}

\subsection{Task 7a}


Each of the three batches of Task 7a were evaluated independently.
The classification performance of the systems were measured using flat and hierarchical evaluation measures \cite{balikas13}. The micro F-measure (MiF) and the Lowest Common Ancestor F-measure (LCA-F)
were used to choose the winners
for each batch \cite{kosmopoulos2015evaluation}.

According to \cite{Demsar06} the appropriate way
to compare multiple classification systems over
multiple datasets is based on their average rank
across all the datasets. On each dataset the system
with the best performance gets rank 1.0, the second best rank 2.0 and so on. In case two
or more systems tie, they all receive the average
rank.
Table \ref{tab:7a_res} presents the average rank (according to
MiF and LCA-F) of each system over all the test
sets for the corresponding batches. Note, that the
average ranks are calculated for the 4 best results
of each system in the batch according to the rules
of the challenge.\newline

\begin{table*}[!htbp]
        \centering
        \begin{tabular}{M{0.4\linewidth}M{0.07\linewidth}M{0.09\linewidth}M{0.07\linewidth}M{0.09\linewidth}M{0.07\linewidth}M{0.09\linewidth}}\hline
        \textbf{System} & \multicolumn{2}{c}{\textbf{Batch 1}} & \multicolumn{2}{c}{\textbf{Batch 2}} & \multicolumn{2}{c}{\textbf{Batch 3}} \\ \hline
        & MiF & LCA-F & MiF & LCA-F & MiF & LCA-F \\ \cline{2-7}
        DeepMeSH5 &	- &- &  1,00 & 1,00	& 1	&1 \\
        DeepMeSH4 &	 - &	 - & 	9,50&	9,50&	2,25&	1,75\\
        DeepMeSH3 &	8,25&	8,50&	3,50&	5,00&	2,5&	2,75\\
        DeepMeSH1 &	5,00&	6,25&	2,00&	2,63&	3,75&	4,13\\
        DeepMeSH2 &	7,25&	7,25&	3,50&	4,50&	4,75&	4,38\\
        MeSHProbeNet-P2 &	2,63&	2,63&	4,63&	5,88&	6,5&	8,25\\
        MeSHProbeNet-P1 &	3,25&	2,13&	6,38&	4,25&	6,88&	6,5\\
        MeSHProbeNet-P3 &	5,00&	4,63&	8,38&	7,25&	7,5&	7,38\\
        MeSHProbeNet-P &	2,38&	3,25&	7,00&	4,38&	8,13&	7,75\\
        MeSHProbeNet-P0 &	1,50&	1,25&	6,25&	5,63&	8,75&	7,88\\
        ceb 1 ensemble 	& - &	 -  &	 - &	 - & 	11&	11\\
        Default MTI &	9,75&	8,75&	12,00&	11,75&	12,25&	12,25\\
        ceb1 &	8,75&	9,25&	11,00&	11,25&	12,25&	13,5\\
        MTI First Line Index &	11,50&	11,25&	13,00&	12,50&	13,25&	12\\
        iria-mix &	 - 	& -  &	14,00&	14,00&	14,5&	14,75\\
        Semantic NoSQL KE 2& 	 - 	& -  &	 - 	& -  &	16&	16\\
        Semantic NoSQL KE 1 &	 - 	& -  &	 - 	& -  &	17&	17,75\\
        \hline
        \end{tabular}
        \caption{Average system ranks across the batches of the Task 7a. A hyphenation symbol (-) is used whenever the system participated in fewer than 4 tests in the
batch. Systems with fewer than 4 participations in all batches are omitted.}\label{tab:7a_res}
\end{table*}

\begin{figure*}[!htb]
\centerline{\includegraphics[width=1\textwidth]{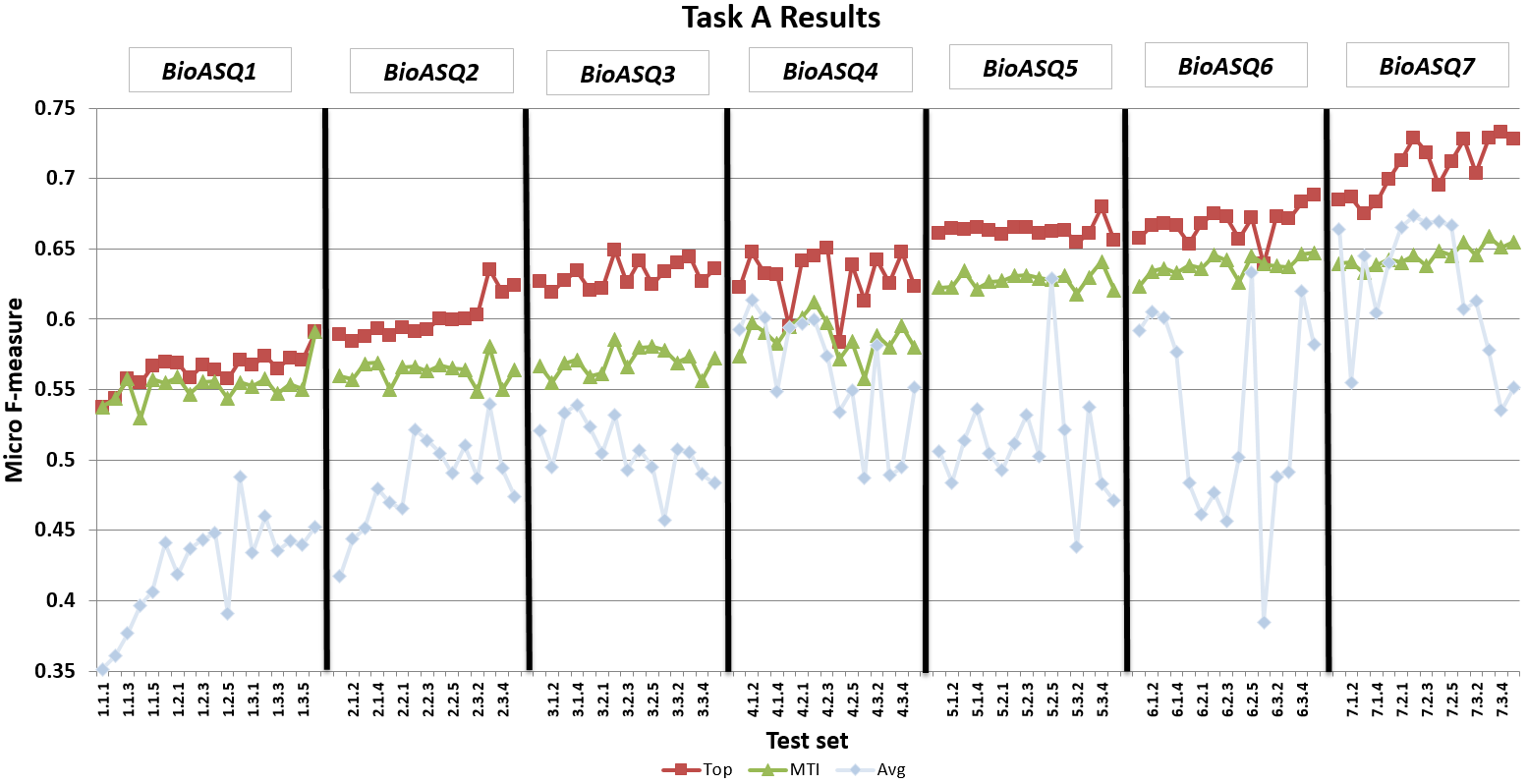}}
\caption{The micro f-measure achieved by systems across different years of the BioASQ challenge. For each test set the micro F-measure is presented for the best performing system (Top) and the MTI, as well as the average micro f-measure of all the participating systems (Avg). }\label{fig:01}
\end{figure*}

\indent The results in Task 7a show that in all test batches and for both flat and hierarchical measures, some systems outperform the strong baselines. In particular, The ``\textit{MeSHProbeNet-P}'' systems achieve the best performance in the first batch,  outperformed by the ``\textit{DeepMeSH}'' systems in the last two batches. More detailed results can be found in the online results page\footnote{\footnotesize \url{http://participants-area.bioasq.org/results/7a/}}. Comparison of these results with corresponding system results from previous years reveals the improvement of both the baseline and the top performing systems through the years of the competition as shown in Figure {\ref{fig:01}}.

\subsection{Task 7b}

\begin{table*}[!htb]
        \centering
        \begin{tabular}{M{0.31\linewidth}M{0.12\linewidth}M{0.12\linewidth}M{0.12\linewidth}M{0.12\linewidth}M{0.12\linewidth}}\hline
        \textbf{System} & \textbf{Mean Precision} & \textbf{Mean Recall} & \textbf{Mean F-measure} & \textbf{MAP} & \textbf{GMAP}  \\ \hline

        aueb-nlp-2 &	0.2060& 	0.4039& 	0.2365& 	\textbf{0.2114}& 	0.0075\\
        aueb-nlp-1 &	0.2124& 	0.4083& 	0.2440& 	0.2086& 	0.0065\\
        aueb-nlp-5 &	\textbf{0.2157}& 	\textbf{0.4235}& 	\textbf{0.2467}& 	0.1821& 	\textbf{0.0098}\\
        MindLab QA Reloaded& 	0.1587 &	0.2760& 	0.1723& 	0.1527& 	0.0013\\
        Deep ML methods for &	0.1331 &	0.2692& 	0.1589&	0.1234& 	0.0009\\
        MindLab Red Lions++ &	0.1371& 	0.2538& 	0.1535& 	0.1187& 	0.0014\\
        aueb-nlp-3 &	0.1488& 	0.3427& 	0.1779& 	0.1149& 	0.0053\\
        MindLab QA System ++& 	0.1288& 	0.2049& 	0.1364& 	0.1136& 	0.0010\\
        aueb-nlp-4 &	0.1520 	&0.3237& 	0.1791& 	0.1116& 	0.0056\\
        MindLab QA System& 	0.1297& 	0.2536& 	0.1478& 	0.1094& 	0.0016\\
        lh\_sys1& 	0.0399 	&0.0810& 	0.0478& 	0.0178& 	0.0001\\
        lh\_sys3 &	0.0233 	&0.0437& 	0.0266& 	0.0151& 	0.0001\\
        lh\_sys5 &	0.0233 	&0.0437& 	0.0266& 	0.0151& 	0.0001\\
        lh\_sys4 &	0.0233 &	0.0437& 	0.0266& 	0.0148& 	0.0001\\
        lh\_sys2 &	0.0182& 	0.0281& 	0.0193& 	0.0051& 	0.0001\\

        \hline
        \end{tabular}
        \caption{Results for snippet retrieval in batch 4 of phase A of Task 7b.}\label{tab:5bA_res_sni}
\end{table*}

\begin{table*}[!htbp]
        \centering
        \begin{tabular}{M{0.27\linewidth}M{0.13\linewidth}M{0.13\linewidth}M{0.13\linewidth}M{0.13\linewidth}M{0.12\linewidth}}\hline
        \textbf{System} & \textbf{Mean Precision} & \textbf{Mean Recall} & \textbf{Mean F-measure} & \textbf{MAP} & \textbf{GMAP}  \\ \hline
      
      aueb-nlp-4& 	0.1750& 	\textbf{0.6266} &	0.2471& 	\textbf{0.1199} &	0.0151\\
      aueb-nlp-2& 	0.1740& 	0.6139& 	0.2449& 	0.1121& 	0.0156\\
      aueb-nlp-5 &	\textbf{0.3599}& 	0.6128& 	\textbf{0.4034} &	0.1102& 	\textbf{0.0164}\\
      aueb-nlp-1 &	0.1700& 	0.5912& 	0.2380& 	0.1041& 	0.0118\\
      auth-qa-1 &	0.2675& 	0.3896& 	0.2894 &	0.1033& 	0.0018\\
      aueb-nlp-3 &	0.1600& 	0.5806& 	0.2266& 	0.0986& 	0.0104\\
      lh\_sys4& 	0.1420 &	0.5490& 	0.2081 &	0.0920& 	0.0069\\
      Ir\_sys1 &	0.1410& 	0.5365& 	0.2059& 	0.0907& 	0.0059\\
      lh\_sys1& 	0.1420 &	0.5449& 	0.2076 &	0.0881& 	0.0063\\
      MindLab QA Reloaded& 	0.1330& 	0.5288& 	0.1950& 	0.0863& 	0.0062\\

        \hline
        \end{tabular}
        \caption{Results for document retrieval in batch 3 of phase A of Task 7b. Only the top-10 systems are presented.}\label{tab:5bA_res_doc}
\end{table*}

\begin{table*}[!htb]
        \centering
        \begin{tabular}
        {M{0.24\linewidth}M{0.0852\linewidth}M{0.0852\linewidth}M{0.0852\linewidth}M{0.0852\linewidth}M{0.0852\linewidth}M{0.0852\linewidth}M{0.0852\linewidth}M{0.0852\linewidth}}
        \hline

        \textbf{System} & \multicolumn{2}{c}{\textbf{Yes/No}} & \multicolumn{3}{c}{\textbf{Factoid}} & \multicolumn{2}{c}{\textbf{List}} \\ 
        \hline
        & Acc. & F1 & Str. Acc. & Len. Acc. & MRR & Prec. & Rec. & F1 \\ \cline{2-9}        
        
        BioBERT-DMIS-3 &	\textbf{0.8286}& 	\textbf{0.8250} &	\textbf{0.2857} &	0.4286 &	0.3452 &	\textbf{0.5653}& 	0.4131& 	\textbf{0.4619}\\
        BioBERT-DMIS &	0.8000& 	0.7822& 	0.2571 &	0.4571& 	0.3224& 	0.5236 &	0.3714 &	0.4202\\
        unipi-quokka-QA-5 &	0.8000 & 	0.7939 &	0.0857 &	0.1714 &	0.1152 &	0.1713 &	\textbf{0.5873} 	&0.2537\\
        BioBERT-DMIS-2 &	0.7429 	&0.7200& 	0.2571 	&0.4571 &	0.3271 &	0.5486 &	0.3992 &	0.4468\\
        BioBERT-DMIS-4 &	0.7429 &	0.7351& 	0.2286& 	0.4571& 	0.3238& 	0.5069 &	0.3575 &	0.4051\\
        google-gold-input-ab &	0.7143 & 	0.6941 &	0.2286 	&0.2857 &	0.2571 &	0.1774 & 	0.4175 &	0.2415\\
        unipi-quokka-QA-4 &	0.7143 & 0.6941& 	0.0857 &	0.1714& 	0.1152 &	0.1713 &	\textbf{0.5873} &	0.2537\\
        unipi-quokka-QA-3 &	0.6857 &	0.6578 &	0.0857 &	0.1714 &	0.1152 &	0.1713 	&\textbf{0.5873 }&	0.2537\\
        google-gold-input &	0.6571 	&	0.6023& 	\textbf{0.2857} &	0.3714 &	0.3167 &	0.2159& 	0.4452& 	0.2824\\
        DMIS &	0.6571 &	0.6023& 	\textbf{0.2857}& 	\textbf{0.5143}& 	\textbf{0.3638}& 	0.5050 &	0.3714 &	0.4124\\
        BioASQ\_Baseline &	0.4857& 	0.4643 &	0.0571 &	0.1429& 	0.0867& 	0.2127 &	0.3619 &	0.2573\\
        \hline
        \end{tabular}
        \caption{Results for batch 5 for exact answers in phase B of Task 7b. Only the top-10 systems are presented along with the BioASQ baseline.
        \label{tab:5bB_res}}
\end{table*}

\textbf{Phase A}: For phase A and for each of the four types of annotations: documents, concepts, snippets and RDF triples, we rank the systems according to the Mean Average Precision (MAP) measure. The final ranking for each batch is
calculated as the average of the individual rankings
in the different categories. In Tables \ref{tab:5bA_res_sni} and \ref{tab:5bA_res_doc}
some indicative results from batches 3 and 4 are presented. Full results are available in the online results page of Task 7b, phase A\footnote{\footnotesize \url{http://participants-area.bioasq.org/results/7b/phaseA/}}. These results are preliminary. The final results for Task 7b, phase A will be available after the manual assessment of the system responses.

\textbf{Phase B}: In phase B of Task 7b the systems were asked to produce exact and ideal answers. For ideal answers, the systems will eventually be ranked according to manual evaluation by the BioASQ experts \cite{balikas13}. Regarding exact answers\footnote{For summary questions, no exact answers are required}, the systems were ranked according to accuracy, F1 score on prediction of yes answer, F1 on prediction of no and macro-averaged F1 score for the yes/no questions, mean reciprocal rank (MRR) for the factoids and mean F-measure for the list questions. Table \ref{tab:5bB_res} shows the results for exact answers for the last batch of Task 7b. These results are preliminary. The full results of phase B of Task 7b are available online\footnote{\footnotesize \url{http://participants-area.bioasq.org/results/7b/phaseB/}}.  The final results for Task 7b, phase B will be available after the manual assessment of the system responses.

\begin{figure*}[!htb]
\centerline{\includegraphics[width=1\textwidth]{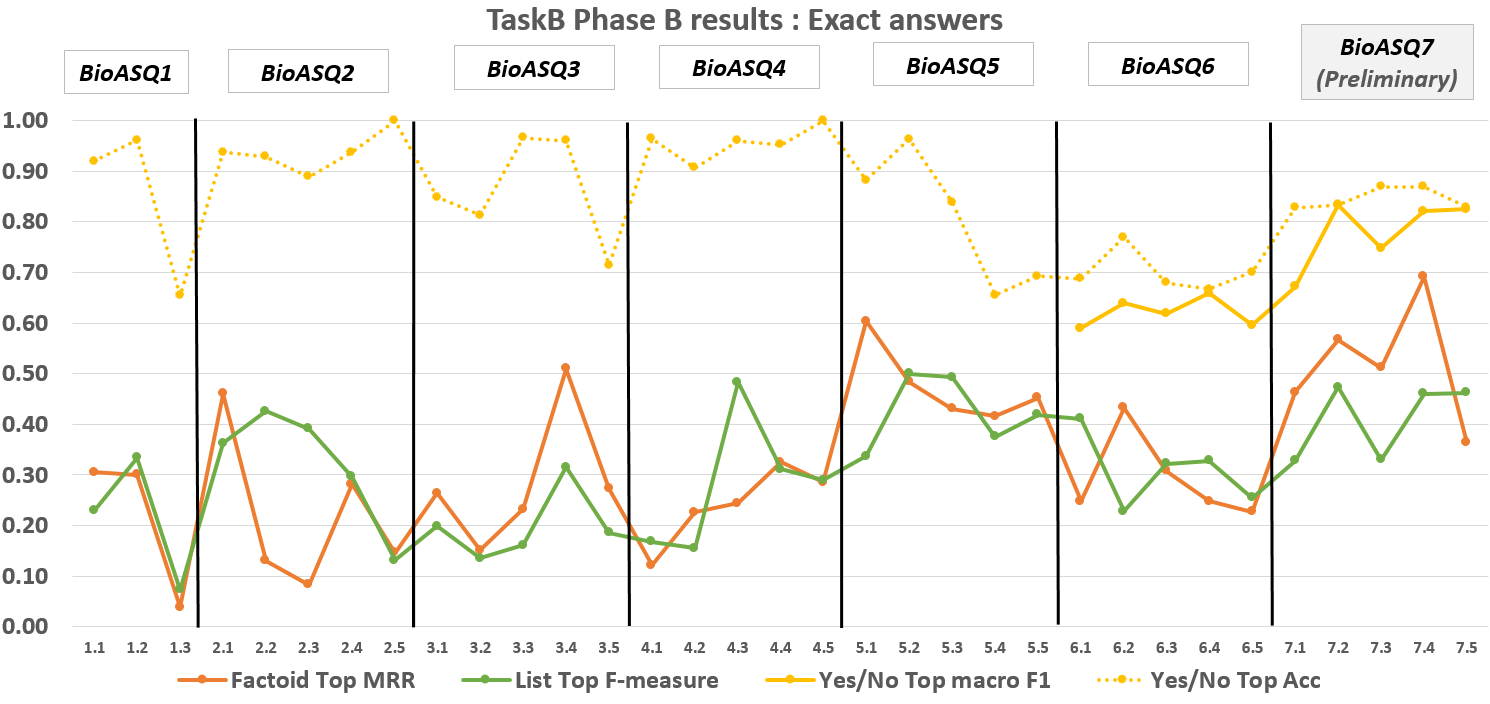}}
\caption{The performance achieved by systems in exact answer generation part of Task B, Phase B, across different years of the BioASQ challenge. For each test set the performance of the best performing system (Top) is presented based on the official evaluation measures. Since BioASQ6 the macro-averaged F1 score (macro F1) is the official measure for Yes/No questions, but accuracy (Acc), the former official measure, is also presented. The results for BioASQ7 are preliminary. The final results for Task 7b, phase B will be available after the manual assessment of the system responses. }\label{fig:02}
\end{figure*}

The results presented in Figure {\ref{fig:02}} show that this year the performance of systems in the yes/no questions, has clearly improved. In batch 5 for example, presented in Table \ref{tab:5bB_res}, some systems outperformed the strong baseline based on previous versions of the OAQA system, with the top system achieving almost double the score of the baseline. Some improvement is also observed in the performance of the top systems for factoid and list questions in the preliminary results. However, there is even more room for improvement in these types of question as can be seen in Figure {\ref{fig:02}}.  

\section{Conclusions}
In this paper, an overview of the seventh BioASQ challenge is presented. The challenge consisted of
two tasks: semantic indexing and question answering. Overall, as in previous years, the best systems were able
to outperform the strong baselines provided by the organizers. This suggests that advances over the
state of the art were achieved through the BioASQ challenge but also that the benchmark in itself is
challenging. Moreover, the shift towards systems that incorporate ideas based on deep learning models observed in the previous year, is even more clear. Novel ideas have been tested and state-of-the-art deep learning methodologies have been adapted to biomedical question answering with great results. Specifically, the breakthroughs in different NLP tasks using clever techniques with the advent of new language-models, such as BERT and gpt-2, gave birth to new approaches that significantly boost the performance of the systems. In the future, we expect novel methodologies, such as the newly proposed XLNet \cite{DBLP:journals/corr/abs-1906-08237}, to further cultivate research in the biomedical information systems field. Consequently, we believe that the challenge is successfully pushing the research frontier  of this domain. In future editions of the challenge, we aim to
provide even more benchmark data derived from a community-driven acquisition process.
\section*{Acknowledgments}

Google was a proud sponsor of the BioASQ Challenge in 2018. The seventh edition of BioASQ is also sponsored by the Atypon
Systems inc.
BioASQ is grateful to NLM for providing
baselines for task 7a and to the CMU
team for providing the baselines for task 7b. Finally,
we would also like to thank all teams for
their participation.\\

%
%
%
\bibliographystyle{splncs04}
\bibliography{BioASQ7_overview.bib}

\end{document}